\documentclass[conference]{IEEEtran}
\IEEEoverridecommandlockouts
\usepackage{cite}
\usepackage{amsmath,amssymb,amsfonts}
\usepackage{algorithmic}
\usepackage{graphicx}
\usepackage{pgfplots}
\usepackage{float}
\usepackage{textcomp}
\usepackage{xcolor}
\usepackage{url}
\usepackage{algorithm2e}
\usepackage{comment}
\def\BibTeX{{\rm B\kern-.05em{\sc i\kern-.025em b}\kern-.08em
    T\kern-.1667em\lower.7ex\hbox{E}\kern-.125emX}}
\pgfplotsset{compat=1.14}
\begin{document}

\title{Hyperparameter Optimisation with Early Termination of Poor Performers}

\author{\IEEEauthorblockN{1\textsuperscript{st} Dobromir Marinov}
\IEEEauthorblockA{\textit{Computer Science and Electric Engineering} \\
\textit{University of Essex}\\
Colchester, UK \\
\url{mr.d.marinov@gmail.com}}
\and
\IEEEauthorblockN{2\textsuperscript{nd} Daniel Karapetyan}
\IEEEauthorblockA{\textit{Computer Science and Electric Engineering} \\
\textit{University of Essex}\\
Colchester, UK \\
\url{daniel.karapetyan@gmail.com}}
}

\maketitle

\begin{abstract}
It is typical for a machine learning system to have numerous hyperparameters that affect its learning rate and prediction quality.
Finding a good combination of the hyperparameters is, however, a challenging job.
This is mainly because evaluation of each combination is extremely expensive computationally; indeed, training a machine learning system on real data with just a single combination of hyperparameters usually takes hours or even days.
In this paper, we address this challenge by trying to predict the performance of the machine learning system with a given combination of hyperparameters without completing the expensive learning process.
Instead, we terminate the training process at an early stage, collect the model performance data and use it to predict which of the combinations of hyperparameters is most promising.
Our preliminary experiments show that such a prediction improves the performance of the commonly used random search approach.
\end{abstract}


\section{Introduction}
\label{sec:intro}

Machine learning algorithms can solve complex problems in a wide range of domains~\cite{langley_applications_1995}.
In some cases, they are even able to achieve better performance than the top human-experts in a given field~\cite{buetti-dinh_deep_2019}.
The performance of machine learning models is determined to a large degree by the appropriate choice of parameters for the algorithms~\cite{probst_tunability:_2018}.
Those parameters are typically called \textit{hyperparameters} and represent values that dictate the training process, such as the learning rate.
The number of hyperparameters can vary quite drastically between algorithms, with some having as little as 3-5 hyperparameters, and others having hundreds of them.
Furthermore, each hyperparameter can take multiple values.
This leads to combinatorial explosion, and the fact that evaluation of each combination of hyperparameters (which includes training) can take significant amount of time, makes the problem particularly challenging.

A recent approach in parameter tuning used within a new optimisation framework, Conditional Markov Chain Search~\cite{KARAPETYAN2017494}, is to predict the performance of a combination of parameters based on information collected after a short run~\cite{KarapetyanParkesStuetzlenst2018}.
This can potentially save computational power, however the use of predictions instead of the actually measured performance reduces the quality of the tuning method.
This paper translates this idea into the machine learning domain, with the aim to prove that the concept is viable and can be further developed to improve existing hyperparameter tuning methods.

In this paper, we do not intend to develop a state-of-the-art hyperparameter tuning method; this is a proof-of-concept project demonstrating that the early prediction of the machine learning system performance can improve the performance of hyperparameter tuning methods.
Further research is needed to develop and generalise the approach and combine it with other hyperparameter tuning approaches.

\section{Related Work}
\emph{Hyperparameter optimisation} is the process of finding the best possible combination of parameters for a given algorithm.
It is represented by the equation:
\begin{equation}
\underset{\lambda \in \Lambda}{\lambda^* = arg\,max\,f(\lambda)} \,,
\label{eq:1}
\end{equation}
where $\Lambda$ is the space of hyperparameter combinations and $f(\lambda)$, $\lambda \in \Lambda$, measures the performance of the machine learning system with the combination of hyperparameters $\lambda$.
The performance is usually defined as the accuracy but any other metric can also be used.
The optimisation algorithms are trying to find the combination of hyperparameters $\lambda = \lambda^*$ that maximises $f(\lambda)$~\cite{feurer_hyperparameter_nodate}.
It is, however, usually impractical to search for the optimal solution $\lambda^*$, and the problem is replaced with searching for a \emph{good} combination of hyperparameters.

The idea behind the commonly used hyperparameter optimisation methods is to restrict the search space in one or another way.
Let $\Lambda' \subset \Lambda$ be a reasonably-sized subset of the space of hyperparameter combinations.
Then we can find the best configuration $\lambda \in \Lambda'$ by exhaustive search.
$\Lambda'$ can be defined statically or acquired dynamically using some intelligent algorithm.

\emph{Grid search} defines $\Lambda'$ in a systematic way, specifically as a grid in $\Lambda$.
While it has the advantage of giving certain guarantees, it has some major drawbacks.
For example, it is not suitable when the number of hyperparameters is large.
Also, it is unreliable in the sense that its performance may significantly depend on how the grid is defined.

\emph{Random search} defines $\Lambda'$ by randomly sampling $\Lambda$.
Similarly to the grid search, this method makes no assumptions about the structure of the space of hyperparameters, however it addresses the drawbacks of the grid search discussed above.
This makes it a popular choice for the hyperparameter optimisation method.



\emph{Bayesian model-based optimisation} constructs $\Lambda'$ dynamically.
Based on the already evaluated hyperparameter combinations, it builds a so-called surrogate model that approximates $f(\lambda)$ function.
Using the surrogate model, the method focuses on the most promising regions of the search space.
In other words, Bayesian model-based optimisation attempts to predict how yet unseen combinations of hyperparameters will perform.
This is different from our approach because our predictions are based on the results of a quick evaluation of a combination of hyperparameters.

\section{Proposed algorithm}

All the hyperparameter optimisation algorithms discussed in Section~\ref{sec:intro} treat the machine learning model as a black box and only take into account the final score that it produces once fully trained, along with the values of the hyperparameters themselves.
The core idea behind the proposed approach is to instead train each model for only a short time and then extract available information to predict the final score of the model.
In other words, following similar work on parameter tuning in optimisation~\cite{KarapetyanParkesStuetzlenst2018}, we hypothesise that the final accuracy of a machine learning model can be predicted from its internal state at the early stages of training.

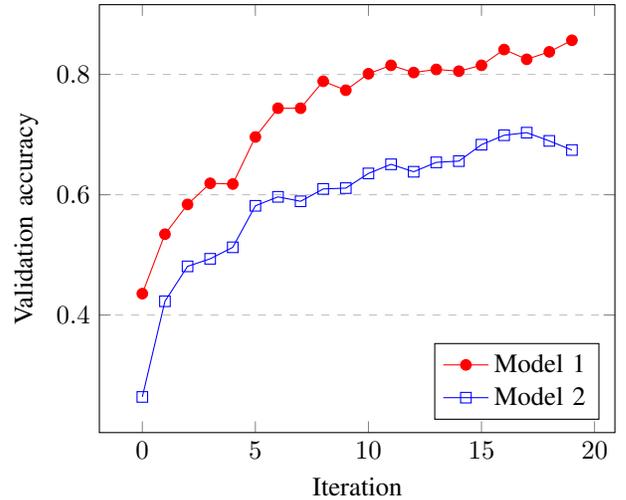
\begin{figure}[tb]
\centering
\begin{tikzpicture}
\begin{axis}
        [
            xlabel={Iteration},
            ylabel={Validation accuracy},
            legend pos=south east,
            ymajorgrids=true,
            grid style=dashed,
            legend cell align=left,
        ]
\addplot[color=red, mark=*]
        coordinates{
            (0,0.4356)
            (1,0.5344)
            (2,0.5839)
            (3,0.619)
            (4,0.6178)
            (5,0.696)
            (6,0.7438)
            (7,0.7438)
            (8,0.7886)
            (9,0.7737)
            (10,0.8011)
            (11,0.8151)
            (12,0.8031)
            (13,0.8083)
            (14,0.8054)
            (15,0.8151)
            (16,0.8413)
            (17,0.8251)
            (18,0.8375)
            (19,0.8567)
        };
\addplot[color=blue, mark=square]
        coordinates{
            (0,0.2639)
            (1,0.4229)
            (2,0.4809)
            (3,0.4935)
            (4,0.5126)
            (5,0.5816)
            (6,0.5965)
            (7,0.5891)
            (8,0.6096)
            (9,0.611)
            (10,0.6354)
            (11,0.6506)
            (12,0.6382)
            (13,0.6538)
            (14,0.6559)
            (15,0.6832)
            (16,0.6989)
            (17,0.7032)
            (18,0.6895)
            (19,0.6743)
        };
\legend{Model 1, Model 2}
\end{axis}
\end{tikzpicture}
\caption{Comparison of the training curves of a high performing model and a low performing one. The difference between the two is present from the start, supporting the theory for using early prediction to identify the better performing model.} 
\label{fig:high_vs_low}
\end{figure}

Consider the example in Figure~\ref{fig:high_vs_low}.
We show here how the validation accuracy of two models changes over time during the training process.
Just by looking at the performance of these models in the first few iterations, we can predict that Model~1 will keep its advantage over Model~2 when the training is complete. 
Hence, by analysing the internal state of a model at the early stages of training, we might be able to predict if the model will perform well after the training is complete.

Based on our observations, we propose a novel approach for hyperparameter optimisation, called \textit{Predictive Hyperparameter Optimisation} (Predictive Hyper-Opt or PHO for short), which creates a pool of partially trained models, predicts the most promising one and evaluates it, in order to find the top performer.
Our algorithm takes as parameters the space $\Lambda'$ of models to explore, the number of models $n$ to fully train at the beginning, the number of iterations $m$ to use for partial training, and the number of models $k$ to be selected, trained and tested at the end of the algorithm.

\begin{enumerate}
    \item
    Randomly select a subset $\Lambda'' \subset \Lambda'$ of size $n$.
    For each $\lambda \in \Lambda''$, fully train the model recording its performance after $m$ iterations and also when the training is complete.

    \item
    Using $\Lambda''$ as the training dataset, train a linear regression $R$ that predicts the performance of a fully trained model based on its performance after $m$ iterations.

    \item
    Train each remaining model, i.e.\ each $\lambda \in \Lambda' \setminus \Lambda''$, to $m$ iterations.
    Record its performance.
    Use regression $R$ to predict the final performance of each $\lambda$.
    
    \item
    Select the top-$k$ models according to their predicted final performance and train each of them fully.
    
    \item
    Of the $n + k$ fully trained models, select and return the one that has the maximum performance.
\end{enumerate}

As a result, the algorithm trains $n + k$ models fully and $|\Lambda'| - n - k$ partially.

The use of linear regression for the predictor of a model's performance is dictated by the size of the training set; any more complex predictor could overfit.
Also, regression gives finer-grain predictions compared to classifiers which is crucial considering the size of the training set and the number of configurations selected based on the predictions.

\section{Computational Experiments}

For our experiments, we used a gradient boosting classifier from the \textit{XGBoost} library~\cite{chen_xgboost:_2016}, trained and validated on the bank marketing dataset~\cite{moro_data-driven_2014}.
We were optimising 5 hyperparameters of the classifier while using default values for the other hyperparameters.
Our search space $\Lambda'$ was obtained as a grid of size $|\Lambda'| = 540$.
The dataset contains data for a marketing campaign with the goal of predicting whether or not a customer will subscribe to a term deposit.
The data includes $45,211$ records and 17 attributes and has been used in related research in hyperparameter optimisation for XGBoost~\cite{Putatunda2018}.
This was randomly split into two subsets, one for training~(77\%) and the other for testing~(33\%).

\begin{figure}[tb]
    \centering
    \begin{tikzpicture}
    \begin{axis}
    [
        ylabel={Model rank},
        xlabel={Final validation accuracy},
        ymajorgrids=true,
        grid style = dashed,
        ymin=0,
        ymax=540,
        x tick label style={
                /pgf/number format/fixed,
                /pgf/number format/fixed zerofill,
                /pgf/number format/precision=3
            }
    ]
    \addplot[const plot, thick, mark=none, color=blue, line width=0.25mm] table [y expr=\thisrowno{0} + 1, x index={1}, col sep=comma] {model_pool_accuracies.csv};
    \end{axis}
    \end{tikzpicture}
    \caption{Cumulative histogram of the accuracies of the models in the pool.
    This demonstrates that the model accuracy significantly depends on the values of hyperparameters, and that there are only a few top-performing models.}
    \label{fig:pool_accuracies}
\end{figure}
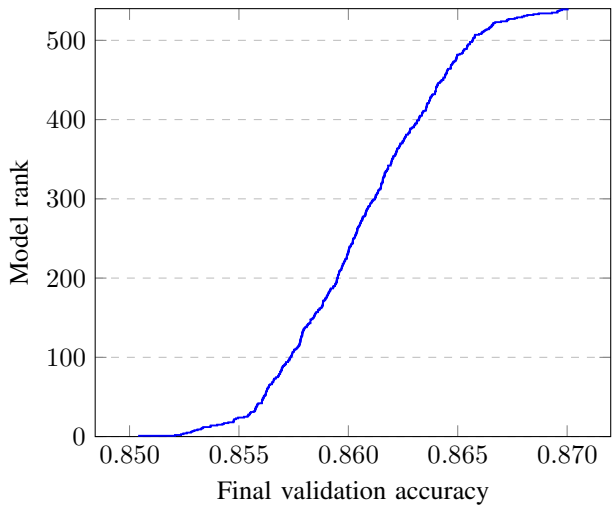

Figure~\ref{fig:pool_accuracies} shows the distribution of performances of all the $\Lambda'$ models.
While the variation along the horizontal axis may seem to be small, a 2\% increase of the accuracy of a machine learning system can have a significant impact in a real world application.
This plot also shows that finding one of the top-performing models is challenging as the majority of the models demonstrate relatively poor performance. 

\begin{figure}[tb]
\centering
\begin{tikzpicture}
\begin{axis}
        [
            xlabel={Accuracy at iteration 2},
            ylabel={Final validation accuracy},
            legend pos=south east,
            ymajorgrids=true,
            grid style=dashed,
        ]
\addplot[blue, thick, only marks, mark options={scale=0.7}] table [x expr=\thisrow{actual}, y expr=\thisrow{iter2}, col sep=comma]{linear_regression.txt};

\addplot[color=purple,  line width=0.25mm, samples at={0.78, 0.85}]{0.17746344745827727 * x + 0.7167788206986645};
\end{axis}
\end{tikzpicture}
\caption{Correlation between the performance at an early epoch and the final accuracy of models. With a correlation coefficient of 0.55, the linear model indicates a moderate positive correlation that supports our hypothesis and holds more predictive capability than choosing models at random.} 
\label{fig:acc_loss}
\end{figure}
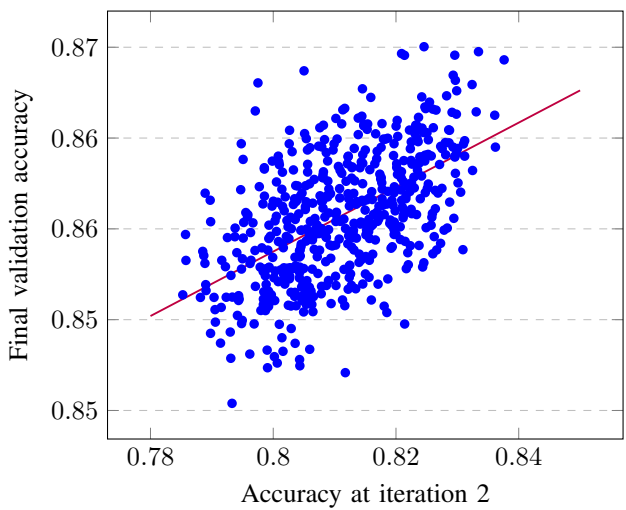

Our second computational experiment verifies the hypothesis that the accuracy of a model can be predicted based on the data collected after just one iteration of training.
Figure~\ref{fig:acc_loss} shows correlation between the accuracy of a model at iteration 2 and its final accuracy when fully trained.
The correlation between these two variables is 0.55, i.e.\ there is a significant potential for exploiting the early signs of a model being successful or unsuccessful.

In our third computational study, we compare the PHO method with the Random Search.
Each experiment consisted of regenerating the split between the training and evaluation datasets and running the PHO and Random Search.
We conducted 5000 such experiments.
As a measure of a model's performance $f(\lambda)$, we used the area under the receiver operating characteristic curve.
The number of the fully trained models was selected as $n = 5$, the number of top models to test $k = 5$ and the number of iterations for a quick training $m = 2$.
To ensure that the running time of the two methods (PHO and Random Search) was equal, PHO was performed first and its overall run-time was recorded. 
Random Search was then given equal time budget.
On average, it performed 57 iterations within this time budget.

All our statistical tests show that PHO performs slightly better than Random Search.
The mean value is 0.0005 higher, and the median value 0.001 higher; the quartiles are also higher.
We also performed two-tailed $t$-test with 0.05 significance level, getting $p = 0.001$, i.e.\ proving that PHO is significantly better than Random Search.

Note that we only study the performance of the hyperparameter optimisation methods and hence there is no need to analyse the performance of the classifier or to compare it against the performance of the other classifiers in the literature.

\section{Conclusions}

In this paper, we discussed a new approach to hyperparameter optimisation for machine learning.
Our method, which we call `Predictive Hyperparameter Optimisation' (PHO), attempts to predict the final performance of a machine learning system based on its early performance during the training process.
A similar approach has been demonstrated to perform well in the domain of discrete optimisation algorithms, and now we translated it into the machine learning domain.

While early termination of training is not a new idea, to the best of our knowledge, all the other methods, such as the Hyperband Algorithm~\cite{li_hyperband:_2017} and the Predictive Termination Criterion~\cite{domhan_speeding_nodate}, do not exploit the knowledge learnt from the terminated models; they all focus on the best configurations.

We performed experiments to compare PHO to pure Random Search; both methods ignore the structure of the search space and hence can be fairly compared.
We observed that the PHO method is slightly better.
This demonstrates the potential of our approach in improving existing or building new hyperparameter optimisation methods.
Further research is needed to prove the concept in other machine learning domains and develop the idea into a robust algorithm, potentially hybridising it with existing intelligent hyperparameter optimisation methods.

\bibliographystyle{bibliography/IEEEtran}
\bibliography{refs}

\end{document}